\documentclass[pmlr]{jmlr}

\usepackage{longtable}

\usepackage{booktabs}
\usepackage[load-configurations=version-1]{siunitx} 

\makeatletter
\def\set@curr@file#1{\def\@curr@file{#1}} 
\makeatother

\theorembodyfont{\upshape}
\theoremheaderfont{\scshape}
\theorempostheader{:}
\theoremsep{\newline}

\usepackage{enumitem}
\usepackage{algorithm}
\usepackage{graphics}
\usepackage{algorithmic}
\usepackage{multirow}
\usepackage{adjustbox}

\title[]{Learning by Passing Tests, with Application to Neural Architecture Search}

\author{\Name{Xuefeng Du\textsuperscript{\dag}}
       \Email{xuefengdu1@gmail.com} 
       \AND
       \Name{Haochen Zhang\textsuperscript{\dag}}
       \Email{zhc12345@mail.ustc.edu.cn} 
       \AND
       \Name{Pengtao Xie\textsuperscript{*}}
       \Email{p1xie@eng.ucsd.edu}\\
       \addr UC San Diego
\AND
       }

\begin{document}

\maketitle

\begin{abstract}
Learning through tests is a broadly used methodology in human learning and shows great effectiveness in improving learning outcome: a sequence of tests are made with increasing levels of difficulty; the learner takes these tests to identify his/her weak points in learning and continuously addresses these weak points to successfully pass these tests. We are interested in investigating whether this powerful learning technique can be borrowed from humans to improve the learning abilities of machines. We propose a novel learning approach called learning by passing tests (LPT). In our approach, a tester model creates increasingly more-difficult tests to evaluate a learner model. The learner tries to continuously improve its learning ability so that it can successfully pass however difficult tests created by the tester. We propose a multi-level optimization framework to formulate LPT, where the tester learns to create difficult and meaningful tests and the learner learns to pass these tests.  We develop an efficient algorithm to solve the LPT problem. Our method is applied for neural architecture search and achieves significant improvement over state-of-the-art baselines on CIFAR-100, CIFAR-10, and ImageNet. 
\end{abstract}

\section{Introduction}

\let\thefootnote\relax\footnotetext{$^\dag$Equal contribution.}

\let\thefootnote\relax\footnotetext{$^*$Corresponding author.}

In human learning, an effective and widely used methodology for improving learning outcome is to let the learner take increasingly more-difficult tests. To successfully pass a more challenging test, the learner needs to gain better learning ability. By progressively passing tests that have increasing levels of difficulty, the learner strengthens his/her learning capability gradually.

\begin{figure}[t]
    \centering
 \includegraphics[width=0.7\columnwidth]{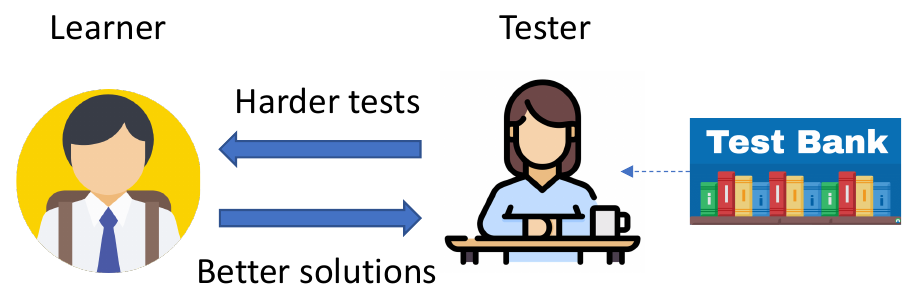}
       \caption{Learning by passing tests. A tester model creates tests with increasing levels of difficulty from a test bank to evaluate a learner model. The learner continuously improves its learning ability to deliver better solutions for passing those difficult tests.}
 \label{fig:illus}
\end{figure}

Inspired by this test-driven learning technique of humans, we are interested in investigating whether this methodology is helpful for improving machine learning as well. We propose a novel machine learning framework called learning by passing tests (LPT). In this framework, there is a ``learner" model and a ``tester" model. The tester creates a sequence of ``tests" with growing levels of difficulty. The learner tries to learn better  so that it can pass these increasingly more-challenging tests. Given a large collection of data examples called ``test bank", the tester   creates a test $T$ by selecting a subset of examples from the test bank. The learner applies its intermediately-trained model $M$ to make predictions on the examples in $T$. The prediction error rate $R$ reflects how difficult this test is. If the learner can make correct predictions on $T$, it means that $T$ is not difficult enough. In this case, the tester will create a more challenging test $T'$ by selecting a new set of examples from the test bank such that the new error rate $R'$ achieved  by $M$ on $T'$ is larger than $R$ achieved on $T$. Given this more demanding test $T'$, the learner re-learns its model to pass $T'$, in a way that the newly-learned model $M'$ achieves a new error rate  $R''$ on $T'$ where $R''$ is smaller than $R'$. This process (as illustrated in Figure~\ref{fig:illus}) iterates until convergence.

In our framework, both the learner and tester perform learning. The learner learns how to best conduct a target task $J_1$ and the tester learns how to create difficult and meaningful tests. To encourage a created test $T$ to be meaningful, the tester trains a model using $T$ to perform a target task $J_2$. If the model performs well on $J_2$, it indicates that $T$ is meaningful. 
The learner has two sets of learnable parameters: neural architecture and network weights. The tester has three learnable modules: data encoder, test creator, and target-task executor. Learning is organized into three stages. In the first stage, the learner trains its network weights on the training set of task $J_1$ with the architecture fixed. In the second stage, the tester trains its data encoder and target-task executor on a created test to perform the target task $J_2$, with the test creator fixed. In the third stage, the learner updates its model architecture by minimizing the predictive loss $L$ on the test created by the tester; the tester updates its test creator by maximizing $L$ and minimizing the loss on the validation set of $J_2$. The three stages are performed jointly end-to-end in a multi-level optimization framework, where different stages influence each other. We apply our method for neural architecture search~\citep{zoph2016neural,liu2018darts,real2019regularized} in image  classification tasks on CIFAR-100, CIFAR-10, and ImageNet~\citep{deng2009imagenet}. Our method achieves significant improvement over state-of-the-art baselines.

The major contributions of this paper are as follows:
\begin{itemize}
\setlength\itemsep{0em}
\item Inspired by the test-driven learning technique of humans, we propose a novel ML approach called learning by passing tests (LPT). In our approach, a tester model creates increasingly more-difficult tests to evaluate a learner model. The learner tries to continuously improve its learning ability so that it can successfully pass however difficult tests created by the tester. 
\item We propose a multi-level optimization framework to formulate LPT where a learner learns to pass tests and a tester learns to create difficult and meaningful tests. 
\item We develop an efficient algorithm to solve LPT. 
\item We apply our approach to neural architecture search and achieve significant improvement on CIFAR-100, CIFAR-10, and ImageNet. 
\end{itemize}

\section{Related Works}
\paragraph{Neural Architecture Search (NAS).} NAS has achieved remarkable progress recently, which aims at searching for  optimal architectures of neural networks to achieve the best predictive performance.  In general, there are three paradigms of methods in NAS: reinforcement learning based  approaches~\citep{zoph2016neural,pham2018efficient,zoph2018learning}, evolutionary algorithm based  approaches~\citep{liu2017hierarchical,real2019regularized}, and differentiable  approaches~\citep{liu2018darts,cai2018proxylessnas,xie2018snas}. 
In RL-based approaches, a policy is learned to iteratively generate new architectures by maximizing a reward which is the accuracy on the validation set. Evolutionary learning approaches represent the architectures as individuals in a population. Individuals with high fitness scores (validation accuracy) have the privilege to generate offspring, which replaces individuals with low fitness scores. Differentiable  approaches adopt a network pruning strategy. On top of an over-parameterized network, the weights of connections between nodes are learned using gradient descent. Then weights close to zero are pruned later on. There have been many efforts devoted to improving differentiable NAS methods. 
 In P-DARTS \citep{chen2019progressive}, the depth of searched architectures is allowed to grow progressively during the training process.   Search space approximation and
regularization approaches are developed to reduce computational overheads and improve search stability.  PC-DARTS \citep{abs-1907-05737} reduces the redundancy in exploring the search space by sampling a small portion  of a super network. Operation search is performed in a subset of channels with the held-out part bypassed in a shortcut. 
Our proposed LPT framework is orthogonal to existing NAS approaches and can be applied to any differentiable NAS methods.

\paragraph{Adversarial Learning.}
Our formulation involves a min-max optimization problem, which is analogous to that in adversarial learning~\citep{goodfellow2014generative} for data generation~\citep{goodfellow2014generative,yu2017seqgan}, domain adaptation~\citep{ganin2015unsupervised}, adversarial attack and defence~\citep{goodfellow2014explaining}, etc. Adversarial learning~\citep{goodfellow2014generative} has been widely applied to 1) data generation~\citep{goodfellow2014generative,yu2017seqgan} where a discriminator tries to distinguish between  generated images and real images and a generator is trained to generate realistic data by making such a discrimination difficult to achieve; 2) domain adaptation~\citep{ganin2015unsupervised} where a discriminator tries to differentiate between source images and target images while the feature learner learns representations which make such a discrimination unachievable; 3) adversarial attack and defence~\citep{goodfellow2014explaining} where an attacker  adds small perturbations to the input data to alter the prediction outcome and the defender trains the model in a way that the prediction outcome remains the same given perturbed inputs. Different from these existing works, in our work, a tester aims to  create harder tests to ``fail" the learner while the learner learns to ``pass" however hard tests created by the tester. \citet{shu2020identifying} proposed to use an adversarial examiner to identify the weakness of a trained model. Our work differs from this work in that we  progressively re-train a learner model based on how it performs on the tests that are  created dynamically by a tester model while the learner model in \citep{shu2020identifying} is fixed and not affected by the examination results. \citet{abs-1912-07768} proposed to learn a generative adversarial network~\citep{goodfellow2014generative} to create synthetic examples which are used to train an NAS model. Our work differs from this work in that we use selected validation examples to validate the model while \citet{abs-1912-07768} use synthesized example to train the model. 
\paragraph{Curriculum Learning.}
Our work is also related to curriculum learning (CL)~\citep{bengio2009curriculum,kumar2010self,jiang2014self,matiisen2019teacher}. In CL, a sequence of training datasets with increasing levels of difficulty is used for model training, from easy to difficult. Our work differs from these previous works in that: our work dynamically selects more-difficult data examples for model evaluation while previous works select data examples for model training.

\begin{figure}[t]
    \centering
 \includegraphics[width=\columnwidth]{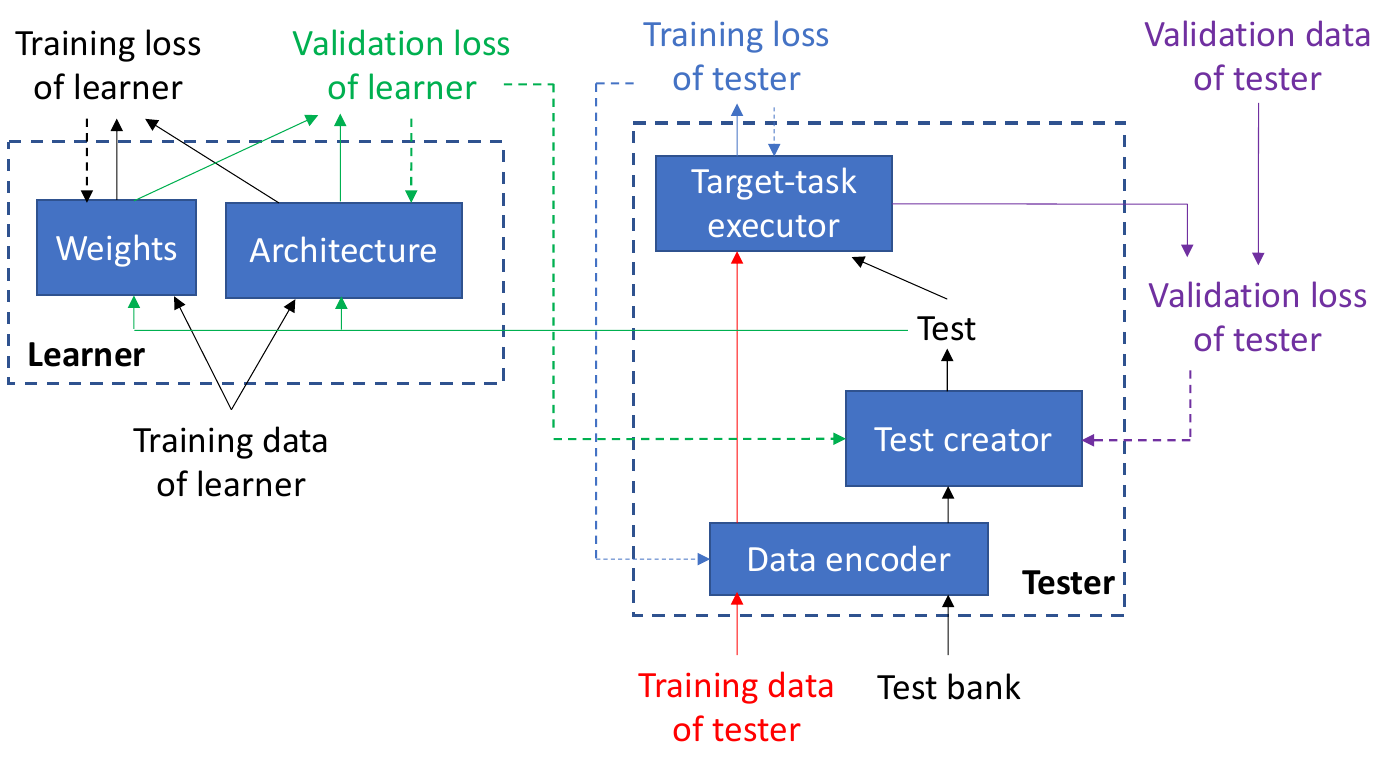}
       \caption{Learning by passing tests. The solid arrows denote the process of making predictions and calculating losses. The dotted arrows denote the process of updating learnable parameters by minimizing corresponding losses.}
 \label{fig:arch}
\end{figure}

\section{Methods}
In this section, we propose a framework to perform learning by passing tests (LPT) (as shown in Figure~\ref{fig:arch}) and develop an optimization algorithm for solving the LPT problem. 
\subsection{Learning by Passing Tests}
In our framework, there is a learner model and a tester model, where the learner studies how to perform a target task $J_1$ such as classification, regression, etc. The eventual goal is to make the learner achieve a better learning outcome with  help from the tester. There is a collection of data examples called ``test bank". The tester creates a test by selecting a subset of examples from the test bank. Given a test $T$, the learner applies its intermediately-trained model $M$ to make predictions on $T$ and measures the prediction error rate $R$. From the perspective of the tester, $R$ indicates how difficult the test $T$ is. If $R$ is small, it means that the learner can easily pass this test. Under such circumstances, the tester will create a more difficult test $T'$ which renders the new error rate $R'$ achieved by $M$ on  $T'$ is larger than $R$. From the learner's perspective, $R'$ indicates how well the learner performs on the test. Given this more difficult test $T'$, the learner refines its model to pass this new test. It aims to learn a new model $M'$ such  that the newer error rate $R''$ achieved by  $M'$ on $T'$ is smaller than $R'$. This process iterates until an equilibrium is reached. In addition to being difficult, the created test should be meaningful as well. It is possible that the test bank contains poor-quality examples where the class labels may be incorrect or the input data instances are outliers. Using an unmeaningful test containing poor-quality examples to guide the learning of the learner may render the learner to overfit these bad-quality examples and generalize poorly on unseen data. To address this problem, we encourage the tester to generate meaningful tests by leveraging the generated tests to perform a target task $J_2$. Specifically, the tester uses  examples in the test to train a model for performing $J_2$. If the performance (e.g., accuracy) $P$ achieved by this model in conducting $J_2$ is high, the test is considered to be meaningful. The tester aims to create a test that can yield a high $P$.

\begin{table}[t]
\caption{Notations in Learning by Passing Tests}
\centering
\begin{tabular}{l|l}
\hline
Notation & Meaning \\
\hline
$A$ & Architecture of the learner\\
$W$ & Network weights of the learner\\
$E$ & Data encoder of the tester\\
$C$ & Test creator of the tester \\
$X$ & Target-task executor of the tester\\
$D_{ln}^{\textrm{(tr)}}$ & Training data of the learner\\
$D_{tt}^{\textrm{(tr)}}$ & Training data of the tester\\
$D_{tt}^{(\textrm{val})}$ & Validation data of the tester\\
$D_b$ & Test bank \\
\hline
\end{tabular}
\label{tb:notations}
\end{table}

In our framework, both the learner and the tester performs learning. The learner studies how to best fulfill the target task $J_1$. The tester studies how to create tests that are difficult and meaningful.  In the learner' model, there are two sets of learnable parameters: model architecture  and network weights. The architecture and weights are both used to make predictions in $J_1$.
The tester's model performs two tasks simultaneously: creating tests and performing another  target-task $J_2$. The model has three learnable modules: data encoder, test creator, and target-task executor, where the test creator performs the task of generating tests and the target-task executor conducts $J_2$.  The test creator and target-task executor share the same data encoder. The data encoder takes a data example $d$ as input and generates a latent representation for this example. Then the representation is fed into the test creator which determines whether $d$ should be selected into the test. The  representation is also fed into the target-task executor which performs prediction on $d$ during performing the target task $J_2$. 

In our framework, the learning of the learner and the tester is organized into three stages. In the first stage, the learner learns its network weights $W$ by minimizing the training loss $L(A, W, D_{ln}^{(\mathrm{tr})})$ defined on the training data $D_{ln}^{(\mathrm{tr})}$ in the  task $J_1$. The architecture $A$ is used to define the training loss, but it is not learned at this stage. If $A$ is learned by minimizing this training loss, a trivial solution will be yielded where $A$ is very large and complex that it can perfectly overfit the training data but will generalize poorly on unseen data. Let $W^*(A)$ denotes the optimally learned $W$ at this stage. 
Note that $W^*$ is a function of $A$ because $W^*$ is a function of the training loss and the training loss is a function of $A$. 
In the second stage, the tester learns its data encoder $E$ and target-task executor $X$ by minimizing the training loss $L(E, X, D_{tt}^{(\mathrm{tr})}) +\gamma L(E, X, \sigma(C, E, D_{b}))$ in the task $J_2$. The training loss consists  of two parts. The first part $L(E, X, D_{tt}^{(\mathrm{tr})})$ is defined on the training dataset $D_{tt}^{\textrm{(tr)}}$ in $J_2$. The second part $L(E, X, \sigma(C, E, D_{b}))$ is defined on the test $\sigma(C, E, D_{b})$ created  by the test creator. To create a test, for each example $d$ in the test bank $D_{b}$, it is first fed into the encoder $E$, then into the test creator $C$, which outputs a binary value  indicating whether $d$ should be selected into the test. $\sigma(C, E, D_{b})$ is the collection of examples whose binary value is equal to 1. $\gamma$ is a tradeoff parameter between these two parts of losses. The creator $C$ is used to define the second-part loss, but it is not learned at this stage. Otherwise, a trivial solution will be yielded where $C$ always sets the binary value to 0 for each test-bank example so that the second-part loss becomes 0. Let $E^*(C)$ and $X^*(C)$ denote the optimally trained $E$ and $X$ at this stage. Note that they are both functions of $C$ since they are functions of the training loss and the training loss is a function of $C$. In the third stage, the learner learns its architecture by trying to pass the test $\sigma(C, E^*(C), D_{b})$ created by the tester. Specifically, the learner aims to minimize its predictive loss   on the test:
\begin{equation}
    L(A, W^{*}(A), \sigma(C, E^{*}(C), D_{b}))=\sum_{d\in \sigma(C, E^{*}(C), D_{b}) } \ell(A, W^{*}(A), d),
\end{equation}
where $d$ is an example in the test and $\ell(A, W^{*}(A), d)$ is the loss defined in this example. 
A smaller $L(A, W^{*}(A), \sigma(C, E^{*}(C), D_{b}))$ indicates that the learner performs well on this test.

Meanwhile, the tester learns its test creator $C$ in a way that $C$ can create a test with more difficulty and meaningfulness. Difficulty is measured by the learner's predictive loss $L(A, W^{*}(A), \sigma(C, E^{*}(C), D_{b}))$ on the test. Given a model $(A, W^{*}(A))$ of the learner and two tests of the same size (same number of examples): $\sigma(C_1, E^{*}(C_1), D_{b})$ created by $C_1$ and $\sigma(C_2, E^{*}(C_2), D_{b})$ created by $C_2$, if $L(A, W^{*}(A), \sigma(C_1, E^{*}(C_1), D_{b}))> L(A, W^{*}(A), \sigma(C_2, E^{*}(C_2), D_{b}))$, it means that $\sigma(C_1, E^{*}(C_1), D_{b})$ is more challenging to pass than $\sigma(C_2, E^{*}(C_2), D_{b})$.   
Therefore, the tester can learn to create a more challenging test by maximizing $L(A, W^{*}(A), \sigma(C, E^{*}(C), D_{b}))$. A trivial solution of increasing $L(A, W^{*}(A), \sigma(C, E^{*}(C), D_{b}))$ is to enlarge the size of the test. But a larger size does not imply more difficulty. To discourage this degenerated solution from happening, we normalize the loss using the size of the test:  
\begin{equation}
    \frac{1}{\left|\sigma\left(C, E^{*}(C), D_{b}\right)\right|} L\left(A, W^{*}\left(A\right), \sigma\left(C, E^{*}(C), D_{b}\right)\right),
\end{equation}
where $|\sigma(C, E^{*}(C), D_{b})|$ is the cardinality of the set $\sigma(C, E^{*}(C), D_{b})$. 
To measure the meaningfulness of a test, we check how well the optimally-trained task executor $X^*(C)$ and data encoder $E^*(C)$ of the tester  perform on the validation data $D_{tt}^{\textrm{(val)}}$ of the target task $J_2$, and the performance is measured by the validation loss: $L(E^{*}(C), X^{*}(C), D_{tt}^{(\mathrm{val})})$. $E^*(C)$ and $X^*(C)$ are trained using the test generated by $C$ in the second stage. If the validation loss is small, it means that the created test is helpful in training the task executor and therefore is considered as being meaningful. To create a meaningful test, the tester learns $C$ by minimizing  $L(E^{*}(C), X^{*}(C), D_{tt}^{(\mathrm{val})})$. In sum, $C$ is learned by maximizing $L(A, W^{*}(A), \sigma(C, E^{*}(C), D_{b}))/|\sigma(C, E^{*}(C), D_{b})|-\lambda L(E^{*}(C), X^{*}(C), D_{tt}^{(\mathrm{val})})$, where $\lambda$ is a tradeoff parameter between these two objectives. 

The three stages are mutually dependent: $W^*(A)$ learned in the first stage and $E^*(C)$ and $X^*(C)$ learned in the second stage are used to define the objective function in the third stage; the updated $C$ and $A$ in the third stage in turn change the objective functions in the first and second stage, which subsequently render $W^*(A)$, $E^*(C)$, and $X^*(C)$ to be changed. Putting these pieces together, we formulate LPT as the following  multi-level optimization problem.
\begin{equation}
    \begin{array}{l}
\max _{C} \min _{A}\;\; \frac{1}{\left|\sigma\left(C, E^{*}(C), D_{b}\right)\right|} L\left(A, W^{*}\left(A\right), \sigma\left(C, E^{*}(C), D_{b}\right)\right)\\ \quad\quad\quad\quad\quad-\lambda L\left(E^{*}(C), X^{*}(C), D_{tt}^{(\mathrm{val})}\right) \textrm{(Stage III)} \\
s.t. \;\;   E^{*}(C), X^{*}(C)=\min _{E, X} \;\; L\left(E, X, D_{tt}^{(\mathrm{tr})}\right) +\gamma L\left(E, X, \sigma\left(C, E, D_{b}\right)\right) \textrm{(II)} \\
\quad\;\;\;
W^{*}\left(A\right)=\min _{W}\;\; L\left(A, W, D_{ln}^{(\mathrm{tr})}\right) \textrm{(Stage I)}
\end{array}
\label{eq:learning_objective}
\end{equation}
This formulation nests three optimization problems. On the constraints of the outer optimization problem are two inner optimization problems corresponding to the first and second learning stage. The objective function of the outer optimization problem corresponds to the third learning stage. 

As of now, the test $\sigma(C, E, D_{b})$ is represented as a subset, which is highly discrete and therefore difficult for optimization. To address this problem, we perform a continuous relaxation of $\sigma(C, E, D_{b})$:
\begin{equation}
   \sigma(C, E, D_{b}) =\{(d,f(d,C,E))|d\in D_{b}\},
\end{equation}
where for each example $d$ in the test bank, the original binary value indicating whether $d$ should be selected is now relaxed to a continuous probability $f(d,C,E)$ representing how likely $d$ should be selected. Under this relaxation, $L(E, X, \sigma(C, E, D_{b}))$ can be computed as follows:
\begin{equation}
    L(E, X, \sigma(C, E, D_{b}))= \sum_{d\in D_{b}} f(d,C,E) \ell (E,X,d),
\end{equation}
where we calculate the loss $\ell (E,X,d)$ on each test-bank example and weigh this loss using $f(d,C,E)$. If $f(d,C,E)$ is small, it means that $d$ is less likely to be selected into the test and its corresponding loss should be down-weighted. Similarly, $L(A, W^{*}(A), \sigma(C, E^{*}(C), D_{b}))$ is calculated as $\sum_{d\in D_{b}} f(d,C,E^{*}(C)) \ell (A, W^{*}(A),d)$. And $|\sigma(C, E^{*}(C), D_{b})|$ can be calculated as
\begin{equation}
    |\sigma(C, E^{*}(C), D_{b})|=\sum_{d\in D_{b}} f(d,C,E^{*}(C)).
\end{equation}
Similar to \citep{liu2018darts}, we represent the architecture $A$ of the learner in a differentiable way. The search space of $A$ is composed of a large number of building blocks. The output of each block is associated with a variable $a$ indicating how important this block is. After learning, blocks whose $a$ is among the largest are retained to form the final architecture. In this end, architecture search amounts to optimizing the set of architecture variables $A=\{a\}$. 

\subsection{Optimization Algorithm}

In this section, we derive an optimization algorithm to solve the LPT problem. Inspired by~\citep{liu2018darts}, we approximate $E^{*}(C)$ and $X^{*}(C)$ using one-step gradient descent update of $E$ and $X$ with respect to $L(E, X, D_{tt}^{(\mathrm{tr})}) +\gamma L(E, X, \sigma(C, E, D_{b}))$ and approximate $W^{*}(A)$ using one-step gradient descent update of $W$  with respect to $L(A, W, D_{ln}^{(\mathrm{tr})})$. Then we plug these approximations into \begin{equation}
\begin{array}{l}
 L(A, W^{*}(A), \sigma(C, E^{*}(C), D_{b}))/|\sigma(C, E^{*}(C), D_{b})|-\lambda L(E^{*}(C), X^{*}(C), D_{tt}^{(\mathrm{val})}),
 \end{array}
 \label{eq:3rd-obj}
\end{equation}
and perform gradient-descent update of $C$ and $A$ with respect to this approximated objective. In the sequel, we use $\nabla^2_{Y,X}f(X,Y)$ to denote $\frac{\partial f(X,Y)}{\partial X\partial Y}$.

Approximating $W^{*}(A)$ using $W'=W - \xi_{ln}  \nabla_{W}L(A, W, D_{ln}^{(\mathrm{tr})})$ where $\xi_{ln}$ is a learning rate and simplifying the notation of $ \sigma(C, E^*(C), D_{b})$ as $\sigma$, we can calculate the approximated gradient of $L\left(A, W^{*}\left(A\right),\sigma\right)$  w.r.t $A$ as:
\begin{equation}
\begin{array}{l}
     \nabla_{A} L\left(A, W^{*}\left(A\right),\sigma\right)\approx  \\
     \nabla_{A}  L\left(A,W - \xi_{ln}  \nabla_{W}L\left(A, W, D_{ln}^{(\mathrm{tr})}\right), \sigma\right)=\\
    \nabla_{A} L\left(A, W^{\prime}, \sigma\right)-\xi_{ln} \nabla_{A, W}^{2} L\left(A, W, D_{ln}^{(\mathrm{tr})}\right) \nabla_{W^{\prime}} L\left(A, W^{\prime},\sigma\right).
\end{array}
\label{eq:descent_arch}
\end{equation}
The second term in the third line   involves expensive matrix-vector product, whose computational complexity can be reduced by a finite difference approximation:
\begin{equation}
\begin{array}{ll}
     \nabla_{A, W}^{2} L\left(A, W, D_{ln}^{(\mathrm{tr})}\right)\nabla_{W^{\prime}} L\left(A, W^{\prime},\sigma\right)\approx
     \frac{1}{2\alpha_{ln}}\left(\nabla_{A} L\left(A, W^{+}, D_{ln}^{(\mathrm{tr})}\right)-\nabla_{A} L\left(A, W^{-}, D_{ln}^{(\mathrm{tr})}\right)\right),
\end{array}
\label{eq:finite-aw}
\end{equation}
where $W^{\pm}=W \pm \alpha_{ln} \nabla_{W^{\prime}} L\left(A, W^{\prime},\sigma\right)$ and $\alpha_{ln}$ is a small scalar that equals $0.01 /\left\|\nabla_{W^{\prime}} L\left(A, W^{\prime},\sigma\right))\right\|_{2}$.
We approximate $E^*(C)$ and $X^*(C)$ using the following one-step gradient descent update of $E$ and $C$ respectively:
\begin{equation}
\begin{array}{l}
    E^{\prime}=E-\xi_{E} \nabla_{E}[ L(E, X, D_{tt}^{(\mathrm{tr})})+\gamma L(E, X, \sigma(C,E,D_b))]\\
    
    X^{\prime}=X-\xi_{X} \nabla_{X}[ L(E, X, D_{tt}^{(\mathrm{tr})})+\gamma L(E, X, \sigma(C,E,D_b))]
    \label{eq:update_ec}
    \end{array}
\end{equation}
where $\xi_{E}$ and $\xi_{X}$ are learning rates. Plugging these approximations into the objective function in Eq.(\ref{eq:3rd-obj}), we can learn $C$ by maximizing the following objective using gradient methods:
\begin{equation}
    L(A, W^{\prime}, \sigma(C, E^{\prime}, D_{b}))/|\sigma(C, E', D_{b})|-\lambda L(E^{\prime}, X^{\prime}, D_{tt}^{(\mathrm{val})})
\end{equation}
The derivative of the second term in this objective with respect to $C$ can be calculated as:
\begin{equation}
\begin{array}{l}
\nabla _{C}L(E^{\prime}, X^{\prime}, D_{tt}^{(\mathrm{val})})=
\frac{\partial E'}{\partial C} \nabla _{E'} L(E^{\prime}, X^{\prime}, D_{tt}^{(\mathrm{val})}) +
\frac{\partial X'}{\partial C}\nabla _{X'} L(E^{\prime}, X^{\prime}, D_{tt}^{(\mathrm{val})})\\
\end{array}
\label{eq:grad_c}
\end{equation}
where 
\begin{equation}
\begin{array}{l}
    \frac{\partial E'}{\partial C}=-\xi_{E}\gamma \nabla^{2}_{C,E} L(E, X, \sigma(C,E,D_b))\\
    \frac{\partial X'}{\partial C}=-\xi_{X}\gamma \nabla^{2}_{C,X} L(E, X, \sigma(C,E,D_b))\\
    \end{array}
    \label{eq:sec-gra-ec}
\end{equation}
Similar to Eq.(\ref{eq:finite-aw}), using finite difference approximation to calculate $\nabla^{2}_{C,E} L(E, X, \sigma(C,E,D_b))$\\$\nabla _{E'} L(E^{\prime}, X^{\prime}, D_{tt}^{(\mathrm{val})})$ and $\nabla^{2}_{C,X} L(E, X, \sigma(C,E,D_b))\nabla _{X'} L(E^{\prime}, X^{\prime}, D_{tt}^{(\mathrm{val})})$, we have:
\begin{equation}
\begin{array}{l}
    \nabla _{C}L(E^{\prime}, X^{\prime}, D_{tt}^{(\mathrm{val})})=\\
    -\gamma\xi_{E}\frac{\nabla_{C}L(E^+,X,\sigma(C,E^+,D_b))-\nabla_{C}L(E^-,X,\sigma(C,E^-,D_b))}{2\alpha_{E}}
    -\gamma\xi_{X}\frac{\nabla_{C}L(E,X^+,\sigma(C,E,D_b))-\nabla_{C}L(E,X^-,\sigma(C,E,D_b))}{2\alpha_{X}}
    \end{array}
\end{equation}
where $E^{\pm}=E\pm\alpha_{E} \nabla_{E^\prime}L(E^\prime,X^\prime,D_{tt}^{\mathrm{(val)}})$ and $X^{\pm}=X\pm\alpha_{X} \nabla_{X^\prime}L(E^\prime,X^\prime,D_{tt}^{\mathrm{(val)}})$. 
For the first term $L(A, W^{\prime}, \sigma(C, E^{\prime}, D_{b}))/|\sigma(C, E', D_{b})|$ in the objective, we can use chain rule to calculate its derivative w.r.t $C$, which involves calculating the derivative of $L(A, W^{\prime}, \sigma(C, E^{\prime}, D_{b}))$ and $|\sigma(C, E', D_{b})|$ w.r.t to $C$. 
The derivative of $L(A, W^{\prime}, \sigma(C, E^{\prime}, D_{b}))$ w.r.t $C$ can be calculated as:
\begin{equation}
\begin{array}{l}
 \nabla _{C}L(A, W^{\prime}, \sigma(C, E^{\prime}, D_{b}))=
 \frac{\partial E'}{\partial C } \nabla_{E^{\prime}}L(A, W^{\prime}, \sigma(C, E^{\prime}, D_{b})),
\end{array}
\label{eq:descent_teach_v2}
\end{equation}
where $ \frac{\partial E'}{\partial C }$ is given in Eq.(\ref{eq:sec-gra-ec}) and $ \nabla^{2}_{C,E} L(E, X, \sigma(C,E,D_b))$
$\times \nabla_{E^{\prime}}L(A, W^{\prime}, \sigma(C, E^{\prime}, D_{b}))$ can be approximated with $\frac{1}{2\alpha_{E}}(\nabla_{C}L(E^+,X,\sigma(C,E^+,D_b))-\nabla_{C}L(E^-,X,\sigma(C,E^-,D_b)))$, where $E^{\pm}$ is $E\pm\alpha_{E}\nabla_{E^{\prime}}L(A, W^{\prime}, \sigma(C, E^{\prime}, D_{b}))$. The derivative of $|\sigma(C, E', D_{b})|=\sum_{d\in D_{b}} f(d,C,E')$ w.r.t $C$ can be calculated as
\begin{equation}
    \sum_{d\in D_{b}} \nabla_C f(d,C,E')+\frac{\partial E'}{\partial C} \nabla_{E'} f(d,C,E')
    \label{eq:grad_cardi}
\end{equation}
where $\frac{\partial E'}{\partial C}$ is given in Eq.(\ref{eq:sec-gra-ec}).  The algorithm for solving LPT is summarized in Algorithm~\ref{algo:algo}.

\begin{algorithm}[h]
\SetAlgoLined
 \While{not converged}{
1. Update the architecture of the  learner  by descending  the gradient calculated in Eq.(\ref{eq:descent_arch})\\
2. Update the test creator of the tester by ascending the gradient calculated in Eq.(\ref{eq:grad_c}-\ref{eq:grad_cardi})\\
3. Update the data encoder and target-task executor of the tester using Eq.(\ref{eq:update_ec})\\
4. Update the network weights of the learner  by descending $\nabla_{W}L(A, W, D_{ln}^{(\mathrm{tr})})$
 }
 \caption{Optimization algorithm for learning by passing tests}
 \label{algo:algo}
\end{algorithm}

\section{Experiments}
We apply LPT for neural architecture search in image classification. Following~\citep{liu2018darts}, we first perform architecture search which finds an optimal cell, then perform architecture evaluation which composes multiple copies of the searched cell into a large network, trains it from scratch, and evaluates the trained model on a test set. We let the target tasks of the learner and that of the tester be the same. Please refer to the supplements for more hyperparameter settings, additional results, and significance tests of results.

\subsection{Datasets}
We used three datasets in the experiments: CIFAR-10, CIFAR-100,  and ImageNet~\citep{deng2009imagenet}. The CIFAR-10 dataset contains 50K training images and 10K testing images, from 10 classes (the number of images in each class is equal). We split the original 50K training set into a 25K training set and a 25K validation set.  
In the sequel, when we mention ``training set", it always refers to the new 25K training set.  
During architecture search, the training set is used as  $D_{ln}^{\textrm{(tr)}}$ of the learner and  $D_{tt}^{\textrm{(tr)}}$ of the tester. The validation set is used as the test bank $D_b$ and the validation data $D_{tt}^{(\textrm{val})}$ of the tester. Under such a setting, the data encoder and target-task executor of the tester are trained on a subset (which is a test) of $D_{tt}^{(\textrm{val})}$ and validated on the entire set of $D_{tt}^{(\textrm{val})}$. The interpretation of doing this is: we select a subset of examples from $D_{tt}^{(\textrm{val})}$ to train a model so that it performs the best on the entire $D_{tt}^{(\textrm{val})}$. During architecture evaluation, the combination of the training data and  validation data is used to train a large network stacking multiple copies of the searched cell. The CIFAR-100 dataset contains 50K training images and 10K testing images, from 100 classes (the number of images in each class is equal). Similar to CIFAR-10, the 50K training images are split into a 25K training set and a 25K validation set. The usage of these subsets is the same as that for CIFAR-10.  
The ImageNet dataset contains a training set of 1.3M images and a validation set of  50K images, from 1000 object classes. The  validation set is used as a test set for architecture evaluation. During architecture search, following~\citep{abs-1907-05737}, 10\% of the 1.3M training images are randomly sampled to form a new training set and another 2.5\% of the 1.3M training images are randomly sampled to form a new architecture validation set. 
The usage of the new training set and architecture validation set is the same as that in CIFAR-10.   
During architecture evaluation, all of the 1.3M training images are used for model training. 
In addition to searching architectures directly on ImageNet data, following~\citep{liu2018darts}, we also evaluate the architectures searched using CIFAR-10 and CIFAR-100 on ImageNet: given a cell searched using CIFAR-10 and CIFAR-100, multiple copies of it compose a large network, which is then trained on the 1.3M training data of ImageNet and evaluated on the 50K test data.
\subsection{Experimental Settings}

Our framework is a general one that can be used together with any differentiable search method. Specifically, we apply our framework to the following NAS methods: 1) DARTS~\citep{liu2018darts}, 2) P-DARTS~\citep{chen2019progressive}, 3) DARTS\textsuperscript{+}~\citep{liang2019darts+}, 4) DARTS\textsuperscript{-}~\citep{abs-2009-01027}, 5) PC-DARTS~\citep{abs-1907-05737}. The search space in these methods are similar. The candidate operations include: $3\times 3$ and $5\times 5$ separable convolutions, $3\times 3$ and $5\times 5$ dilated separable convolutions, $3\times 3$ max pooling, $3\times 3$ average pooling, identity, and zero. In LPT, the network of the learner is a stack of multiple cells, each consisting of 7 nodes. For the data encoder of the tester, we tried ResNet-18 and ResNet-50~\citep{resnet}. For the test creator and target-task executor, they are set to one feed-forward layer. $\lambda$ and $\gamma$ are tuned using a  5k held-out dataset in $\{0.1,0.5,1,2,3\}$.  In most experiments, $\lambda$ and $\gamma$ are  set to 1 except for P-DARTS and PC-DARTS. For P-DARTS,  $\lambda,\gamma$ are set to $0.5,1$ for CIFAR-10 and $1,0.5$ for CIFAR-100. For PC-DARTS, we use $\lambda=3,\gamma=1$ and $\lambda=0.1,\gamma=1$ for CIFAR-10 and CIFAR-100, respectively.

For CIFAR-10 and CIFAR-100, during architecture search, the learner's network is a stack of 8 cells, with the initial channel number set to 16. The search is performed for 50 epochs, with a batch size of 64. The hyperparameters for the learner's architecture and weights are set in the same way as DARTS, P-DARTS, DARTS\textsuperscript{+}, and DARTS\textsuperscript{-}. The data encoder and target-task executor of the tester are optimized using SGD with a momentum of 0.9 and a weight decay of 3e-4. The initial learning rate is set to 0.025 with a cosine decay scheduler. The test creator is optimized with the Adam optimizer~\citep{adam} with a learning rate of 3e-4 and a weight decay of 1e-3. During architecture evaluation, 20 copies of the searched cell are stacked to form the learner's network, with the initial channel number set to 36. The network is trained for 600 epochs with a batch size of 96 (for both CIFAR-10 and CIFAR-100). The experiments are performed on a single Tesla v100. For ImageNet, 
following~\citep{liu2018darts}, we take the architecture searched on CIFAR-10 and evaluate it on ImageNet. 
We stack 14 cells (searched on CIFAR-10) to form a large network and set the initial channel number as 48. The network is trained for 250 epochs with a batch size of 1024 on 8 Tesla v100s. Each experiment on LPT is repeated for ten times with the random seed to be from 1 to 10. We report the mean and standard deviation of results obtained from the 10 runs.

\subsection{Results}

\begin{table}[t]
    \caption{ Results on CIFAR-100, including classification error (\%) on the test set, number of parameters (millions) in the searched architecture, and search cost (GPU days). 
    LPT-R18-DARTS-1st denotes that our method LPT is applied to the search space of DARTS. Similar meanings hold for other notations in such a format. R18 and R50 denote that the data encoder of the tester in LPT is set to ResNet-18 and ResNet-50 respectively.  DARTS-1st  and 
DARTS-2nd denotes that first order and second order approximation is used in DARTS. 
    * means the results are taken from DARTS$^{-}$ \citep{abs-2009-01027}. $\dag$ means we re-ran this method for 10 times. $\Delta$ means the algorithm ran for 600 epochs instead of 2000 epochs in the architecture evaluation stage, to ensure a fair comparison with other methods (where the epoch number is 600). The search cost is measured by GPU days on a Tesla v100. 
    }
        \centering
    \begin{tabular}{l|ccc}
    \toprule
    Method & Error(\%)& Param(M)& Cost\\
    \midrule
    *ResNet \citep{he2016deep}&22.10&1.7&-\\
     *DenseNet \citep{HuangLMW17}&17.18&25.6 &-\\
    \hline
    *PNAS \citep{LiuZNSHLFYHM18}&19.53&3.2&150\\
    *ENAS \citep{pham2018efficient}&19.43&4.6&0.5\\
        *AmoebaNet \citep{real2019regularized}&18.93&3.1&3150\\
    \hline
    *GDAS \citep{DongY19}&18.38&3.4&0.2\\ 
    *R-DARTS \citep{ZelaESMBH20}&18.01$\pm$0.26&-&1.6
    \\
      *DropNAS \citep{HongL0TWL020} & 16.39&4.4&0.7 \\
\hline
\hline
     ${}^{\dag}$DARTS-1st \citep{liu2018darts}  &20.52$\pm$0.31 &1.8 &0.4\\
     $\;\;$LPT-R18-DARTS-1st (ours) &\textbf{19.11}$\pm$0.11&2.1&0.6 \\ 
     \hline
            *DARTS-2nd \citep{liu2018darts}  & 20.58$\pm$0.44&1.8&1.5 \\
             $\;\;$LPT-R18-DARTS-2nd (ours) &19.47$\pm$0.20 & 2.1&1.8 \\
               $\;\;$LPT-R50-DARTS-2nd (ours) &\textbf{18.40}$\pm$0.16 &2.5&2.0 \\
            \hline
      *DARTS$^{-}$ \citep{abs-2009-01027}&17.51$\pm$0.25&3.3&0.4\\
      ${}^{\dag}$DARTS$^{-}$ \citep{abs-2009-01027}& 18.97$\pm$0.16& 3.1&0.4\\
           $\;\;$LPT-R18-DARTS$^{-}$ (ours) &18.28$\pm$0.14&3.4& 0.6\\
     \hline
     ${}^{\Delta}$DARTS$^{+}$ \citep{abs-1909-06035}&17.11$\pm$0.43&3.8&0.2\\
                      $\;\;$LPT-R18-DARTS$^{+}$ (ours) &\textbf{16.58}$\pm$0.19& 3.7&0.3 \\
        \hline
      $\dag$PC-DARTS \citep{abs-1907-05737} &17.96$\pm$0.15&3.9&0.1 \\
       $\;\;$LPT-R18-PC-DARTS (ours)&17.04$\pm$0.05&3.6&0.1 \\
         $\;\;$LPT-R50-PC-DARTS (ours)& \textbf{16.97}$\pm$0.21&4.0 &0.1 \\
        \hline
           *P-DARTS \citep{chen2019progressive}&17.49&3.6&0.3\\ 
       $\;\;$LPT-R18-P-DARTS (ours) &\textbf{16.28$\pm$0.10}&3.8& 0.5\\
        $\;\;$LPT-R50-P-DARTS (ours) & 16.38$\pm$0.07& 3.6& 0.5 \\
        \bottomrule
    \end{tabular} 
    \label{tab:cifar100}
\end{table}

\begin{table}[t]
\caption{
    Results on CIFAR-10. 
    * means the results are taken from DARTS$^{-}$ \citep{abs-2009-01027}, NoisyDARTS \citep{abs-2005-03566},  and DrNAS \citep{abs-2006-10355}.
    The rest notations are the same as those in Table~\ref{tab:cifar100}. 
    }
        \label{tab:cifar10}
    \centering

    \begin{tabular}{l|ccc}
    \toprule
    Method& Error(\%)& Param(M) & Cost\\
    \midrule 
    *DenseNet
    \citep{HuangLMW17}&3.46&25.6 &-\\
    \hline
     *HierEvol \citep{liu2017hierarchical}&3.75$\pm$0.12& 15.7 &300\\
    *NAONet-WS \citep{LuoTQCL18} & 3.53 & 3.1&0.4 \\
        *PNAS \citep{LiuZNSHLFYHM18} &3.41$\pm$0.09  &3.2& 225\\
        *ENAS \citep{pham2018efficient} &2.89 & 4.6  &0.5 \\
    *NASNet-A \citep{zoph2018learning} & 2.65 & 3.3& 1800\\
    *AmoebaNet-B \citep{real2019regularized} & 2.55$\pm$0.05 & 2.8&3150  \\
    \hline
        *R-DARTS \citep{ZelaESMBH20} &2.95$\pm$0.21  &- & 1.6 \\
            *GDAS \citep{DongY19}&2.93& 3.4& 0.2 \\
 *GTN~\citep{abs-1912-07768}& 2.92$\pm$0.06 & 8.2&  0.67\\
    *SNAS \citep{xie2018snas} &2.85 & 2.8& 1.5\\
        *BayesNAS \citep{ZhouYWP19} &2.81$\pm$0.04 &3.4&0.2 \\
        *MergeNAS \citep{WangXYYHS20} &2.73$\pm$0.02 &2.9 & 0.2 \\
        *NoisyDARTS \citep{abs-2005-03566} &2.70$\pm$0.23&3.3  & 0.4 \\
            *ASAP \citep{NoyNRZDFGZ20} &2.68$\pm$0.11 & 2.5&0.2 \\
                *SDARTS
    \citep{abs-2002-05283}&2.61$\pm$0.02 & 3.3& 1.3 \\
            *DropNAS \citep{HongL0TWL020} &2.58$\pm$0.14 & 4.1&0.6 \\
    *FairDARTS \citep{abs-1911-12126} &2.54 &3.3 &0.4 \\
       *DrNAS \citep{abs-2006-10355} &2.54$\pm$0.03&4.0&  0.4\\
    \hline
        \hline
            *DARTS-1st \citep{liu2018darts} &3.00$\pm$0.14&3.3&  0.4\\
        $\;\;$LPT-R18-DARTS-1st (ours) &\textbf{2.85}$\pm$0.09 &2.7&0.6 \\
        \hline
               *DARTS-2nd \citep{liu2018darts} &2.76$\pm$0.09&3.3&  1.5\\
           $\;\;$LPT-R18-DARTS-2nd (ours)  &2.72$\pm$0.07&3.4& 1.8 \\
            $\;\;$LPT-R50-DARTS-2nd (ours) &  \textbf{2.68}$\pm$0.02  &3.4& 2.0\\
            \hline
             *DARTS$^{-}$ \citep{abs-2009-01027}&2.59$\pm$0.08&  3.5&0.4\\
             ${}^{\dag}$DARTS$^{-}$ \citep{abs-2009-01027}& 2.97$\pm$0.04& 3.3&0.4\\
         $\;\;$LPT-R18-DARTS$^{-}$ (ours) &2.74$\pm$0.07&3.4& 0.6\\
         \hline
             ${}^{\Delta}$DARTS$^{+}$ \citep{abs-1909-06035}&2.83$\pm$0.05&3.7&0.4\\
           $\;\;$LPT-R18-DARTS$^{+}$ (ours) &\textbf{2.69}$\pm$0.05&3.6& 0.5\\
     \hline
      *PC-DARTS \citep{abs-1907-05737} &\textbf{2.57}$\pm$0.07&3.6& 0.1\\
       $\;\;$LPT-R18-PC-DARTS (ours)& 2.65$\pm$0.17&3.7&0.1\\
     \hline
    *P-DARTS \citep{chen2019progressive}& 2.50&3.4&  0.3\\
     $\;\;$LPT-R18-P-DARTS (ours)& 2.58$\pm$0.14& 3.3 & 0.5 \\
        \bottomrule
    \end{tabular}
\end{table}

\begin{table*}[!ht]
   \caption{Results on ImageNet, including top-1 and top-5 classification errors on the test set, number of weight parameters (millions), and search cost (GPU days). * means the results are taken from DARTS$^{-}$ \citep{abs-2009-01027} and DrNAS \citep{abs-2006-10355}. The rest notations are the same as those in Table 2 in the main paper.  
    The first row block shows networks designed by human manually. 
    The second row block shows non-gradient based search methods. 
    The third block shows gradient-based methods. 
    $\ddag$ means the results following the hyperparameters selected for CIFAR10/100. The hyperparameter for CIFAR100 is used when directly searching on ImageNet. 
    }
    \centering
        \begin{adjustbox}{width=\columnwidth,center}
    \begin{tabular}{l|cccc}
    \toprule
  \multirow{ 2}{*}{Method}   & Top-1  &Top-5 &Param & Cost \\
         & Error (\%) & Error (\%)&(M) & (GPU days)\\
    \midrule
    *Inception-v1 \citep{googlenet}&30.2 &10.1&6.6&- \\
    *MobileNet \citep{HowardZCKWWAA17} &  29.4& 10.5 &4.2&- \\
    *ShuffleNet 2$\times$ (v1) \citep{ZhangZLS18} &  26.4 &10.2 & 5.4&-\\
    *ShuffleNet 2$\times$ (v2) \citep{MaZZS18} &  25.1 &7.6 & 7.4&-\\
    \hline
    *NASNet-A \citep{zoph2018learning} &26.0 &8.4 &5.3 &1800 \\
    *PNAS \citep{LiuZNSHLFYHM18} &25.8 &8.1  &5.1 &225 \\
    *MnasNet-92 \citep{TanCPVSHL19} & 25.2 & 8.0& 4.4&1667\\
        *AmoebaNet-C \citep{real2019regularized} &  24.3 &7.6 &6.4&3150 \\
    \hline
     *SNAS-CIFAR10 \citep{xie2018snas} & 27.3 &9.2 &4.3 &1.5 \\
          *BayesNAS-CIFAR10 \citep{ZhouYWP19} &26.5 &8.9 &3.9&0.2 \\
                    *PARSEC-CIFAR10 \citep{abs-1902-05116} & 26.0 &8.4&5.6&1.0 \\
     *GDAS-CIFAR10 \citep{DongY19} &  26.0&8.5 &5.3 & 0.2\\
                 *DSNAS-ImageNet \citep{HuXZLSLL20} &25.7& 8.1 &- & -\\
          *SDARTS-ADV-CIFAR10 \citep{abs-2002-05283}&25.2& 7.8 &5.4& 1.3 \\
           *PC-DARTS-CIFAR10 \citep{abs-1907-05737} & 25.1 &7.8&5.3&0.1\\
                *ProxylessNAS-ImageNet \citep{cai2018proxylessnas} & 24.9 &7.5 &7.1 &8.3  \\
          *FairDARTS-CIFAR10 \citep{abs-1911-12126} &24.9 &7.5 &4.8 &0.4 \\
     *FairDARTS-ImageNet \citep{abs-1911-12126} &24.4 &7.4 &4.3 &3.0 \\
             *DrNAS-ImageNet \citep{abs-2006-10355} & 24.2 &7.3& 5.2&3.9\\
         *DARTS$^{+}$-ImageNet \citep{abs-1909-06035}& 23.9& 7.4&5.1&6.8\\ 
        *DARTS$^{-}$-ImageNet \citep{abs-2009-01027}&23.8& 7.0&4.9&4.5\\
     *DARTS$^{+}$-CIFAR100 \citep{abs-1909-06035}&23.7& 7.2&5.1&0.2\\
     \hline 
       \hline
            *DARTS-2nd-CIFAR10 \citep{liu2018darts}  & 26.7 &8.7&4.7&1.5 \\
        $\;\;$LPT-R18-DARTS-2nd-CIFAR10 (ours) & 25.3&7.9&4.7&1.8 \\ 
        \hline
          *P-DARTS (CIFAR10) \citep{chen2019progressive}&24.4 &7.4&4.9&0.3\\
        $\ddag$LPT-R18-P-DARTS-CIFAR10 (ours) & 24.2&  7.3&4.9&0.5 \\
        \hline
             *P-DARTS (CIFAR100) \citep{chen2019progressive}&24.7& 7.5&5.1&0.3\\
           $\ddag$LPT-R18-P-DARTS-CIFAR100 (ours) & 24.0& 7.1&5.3&0.5\\
           \hline
            *PC-DARTS-ImageNet \citep{abs-1907-05737} &  24.2 &7.3&5.3&3.8\\
               $\ddag$LPT-R18-PC-DARTS-ImageNet (ours)& \textbf{23.4} &  \textbf{6.8}&5.7&4.0\\ 
        \bottomrule
    \end{tabular}
    \end{adjustbox}
    \label{tab:imagenet}
\end{table*}

Table~\ref{tab:cifar100} shows the classification error (\%), number of weight parameters (millions), and search cost (GPU days) of different NAS methods on CIFAR-100. From this table, we make the following observations. \textbf{First}, when our method LPT is applied to different NAS baselines including DARTS-1st (first order approximation), DARTS-2nd (second order approximation), DARTS$^{-}$ (our run), DARTS$^{+}$, PC-DARTS, and P-DARTS, the classification errors of these baselines can be significantly reduced. For example, applying our method to P-DARTS, the error reduces from 17.49\% to 16.28\%. Applying our method to DARTS-2nd, the error reduces from 20.58\% to 18.40\%. This demonstrates the effectiveness of our method in searching for a better architecture. In our method, the learner continuously improves its  architecture by passing the tests created by the tester with increasing levels of difficulty. These tests can help the learner to identify the weakness of its architecture and provide guidance on how to improve it. Our method creates a new test on the fly based on how the learner performs in the previous round. From the test bank, the tester selects a subset of difficult examples to evaluate the learner. This new test poses a greater challenge to the learner and encourages the learner to improve its architecture so that it can overcome the new challenge. In contrast, in baseline NAS approaches, a single fixed validation set is used to evaluate the learner. The learner can achieve a good performance via ``cheating": focusing on performing well on the majority of easy examples and ignoring the minority of difficult examples. As a result, the learner's architecture does not have the ability to deal with challenging cases in the unseen data. \textbf{Second}, LPT-R50-DARTS-2nd outperforms LPT-R18-DARTS-2nd, where the former uses ResNet-50 as the data encoder in the tester while the latter uses ResNet-18.  ResNet-50 has a better ability of learning representations than ResNet-18 since it is ``deeper": 50 layers versus 18 layers. 
This shows that a ``stronger" tester can help the learner to learn better. With a more powerful data encoder, the tester can better understand examples in the test bank and can make better decisions in creating difficult and meaningful tests. Tests with better quality can  evaluate the learner more effectively  and help to improve the learner's learning capability. When our method is applied to PC-DARTS and P-DARTS, the performance difference resulting from ResNet-18 and  ResNet-50 is not statistically significant. 
\textbf{Third}, our method LPT-R18-P-DARTS  achieves the best performance among all methods, which further demonstrates the effectiveness of LPT in driving the frontiers of neural architecture search forward. \textbf{Fourth}, the number of weight parameters and search costs corresponding to our methods are on par with those in differentiable NAS baselines. This shows that LPT is able to search better-performing architectures without significantly increasing network size and search cost.  A few additional remarks: 1) On CIFAR-100, DARTS-2nd with second-order approximation in the optimization algorithm is not advantageous compared with DARTS-1st which uses first-order approximation; 2) In our run of DARTS$^{-}$, we were not able to achieve the performance reported in~\citep{abs-2009-01027}; 3) In our run of DARTS$^+$,  in the architecture evaluation stage, we set the number of epochs to 600  instead of 2000 as  in~\citep{abs-1909-06035},  to ensure a fair comparison with other methods (where the epoch number is 600).

Table~\ref{tab:cifar10} shows the classification error (\%), number of weight parameters (millions), and search cost (GPU days) of different NAS methods on CIFAR-10. As can be seen, applying our proposed LPT to DARTS-1st, DARTS-2nd, DARTS$^{-}$ (our run), and DARTS$^{+}$ significantly reduces the errors of these baselines. 
For example, with the usage of LPT, the error of DARTS-2nd is reduced from 2.76\% to 2.68\%.  This further demonstrates the efficacy of our  method in searching better-performing architectures, by creating tests  with increasing levels of difficulty and improving the learner through taking these tests. On PC-DARTS and P-DARTS, applying our method does not yield better performance.

Table~\ref{tab:imagenet} shows the results on ImageNet, including top-1 and top-5 classification errors on the test set. In our proposed LPT-R18-PC-DARTS-ImageNet, the architecture is searched on ImageNet, where our method performs much better than PC-DARTS-ImageNet and achieves the lowest error (23.4\% top-1 error and 6.8\% top-5 error) among all methods in Table~\ref{tab:imagenet}. In our methods including  LPT-R18-P-DARTS-CIFAR100, LPT-R18-P-DARTS-CIFAR10, and LPT-R18-DARTS-2nd-CIFAR10, the architectures are searched on CIFAR-10 or CIFAR-100 and evaluated on ImageNet, where these methods outperform their corresponding baselines P-DARTS-CIFAR100, P-DARTS-CIFAR10,  and DARTS-2nd-CIFAR10. 
These results further demonstrate the effectiveness of our method.

\subsection{Ablation Studies}
In order to evaluate the effectiveness of individual modules in LPT, we compare the full LPT framework with the following ablation settings.
\begin{itemize}[leftmargin=*]
    \item \textbf{Ablation setting 1}. In this setting, the tester creates tests solely by maximizing their level of difficulty, without considering their meaningfulness. Accordingly, the second stage in LPT where the tester learns to perform a target-task by leveraging the created tests is removed. 
    The tester 
    directly learns a selection scalar $s(d)\in[0,1]$ for each  example $d$ in the test bank  without going through a data encoder or test creator. The corresponding formulation is:
        \begin{equation}
    \begin{array}{l}
\max _{S} \min _{A} \;\; \frac{1}{\sum_{d\in D_{b}}s(d)} \sum_{d\in D_{b}} s(d) \ell (A, W^{*}(A),d)\\
s.t. \;\;    W^{*}(A)=\min _{W} \;\; L\left(A, W, D_{ln}^{(\mathrm{tr})}\right)
\end{array}
\end{equation}
where $S=\{s(d)|d\in D_{b}\}$. In this study, $\lambda$ and $\gamma$ are both set to 1. The data encoder of the tester is ResNet-18. For CIFAR-100, 
LPT is applied to P-DARTS and DARTS-2nd. 
For CIFAR-10, LPT is applied to DARTS-2nd. 
  \item \textbf{Ablation setting 2}. In this setting, in the second stage of LPT, the tester is trained solely based on the created test, without using the training data of the target task. 
  The corresponding formulation is: 
    \begin{equation}
    \begin{array}{l}
\max _{C} \min _{A} \;\; \frac{1}{\left|\sigma\left(C, E^{*}(C), D_{b}\right)\right|} L\left(A, W^{*}\left(A\right), \sigma\left(C, E^{*}(C), D_{b}\right)\right)\\ \quad\quad\quad\quad\quad-\lambda L\left(E^{*}(C), X^{*}(C), D_{tt}^{(\mathrm{val})}\right) \\
s.t. \;\;   E^{*}(C), X^{*}(C)=\min _{E, X} \;\; L\left(E, X, \sigma\left(C, E, D_{b}\right)\right) \\
\quad\;\;\; W^{*}\left(A\right)=\min _{W}\;\;L\left(A, W, D_{ln}^{(\mathrm{tr})}\right)
\end{array}
\end{equation}
In this study,  $\lambda$ and $\gamma$ are both set to 1. The data encoder of the tester is ResNet-18. For CIFAR-100, 
LPT is applied to P-DARTS and DARTS-2nd. 
For CIFAR-10, LPT is applied to DARTS-2nd.
    \item Ablation study on $\lambda$. We are interested in how the learner's performance varies as the tradeoff parameter $\lambda$ in Eq.(\ref{eq:learning_objective}) increases. In this study, the other tradeoff parameter $\gamma$ in Eq.(\ref{eq:learning_objective}) is set to 1. For both CIFAR-100 and CIFAR-10, we randomly sample 5K data from the 25K training and 25K validation data, and use it as a test set to report performance in this ablation study. 
    The rest 45K data is used as before. 
    Tester's data encoder is ResNe-18. LPT is applied to P-DARTS. 
    \item Ablation study on $\gamma$. We investigate how the learner's performance varies as c increases. In this study, the other tradeoff parameter $\lambda$ is set to 1. Similar to the ablation study on $\lambda$, on 5K randomly-sampled test data, we report performance of architectures searched under different values of $\gamma$. 
    Tester's data encoder is ResNe-18. LPT is applied to P-DARTS. 
\end{itemize}

\begin{table}[t]
\caption{Results for ablation setting 1. ``Difficult only" denotes that the tester creates tests solely by maximizing their level of difficulty,  without considering their meaningfulness, i.e., the tester does not use the tests for learning to perform the target task. ``Difficult + meaningful" denotes the full LPT framework where the tester creates tests by maximizing both  difficulty and meaningfulness. }
    \centering
    \begin{tabular}{l|c}
    \hline
    Method & Error (\%)\\
    \hline
     Difficult only (DARTS-2nd, CIFAR-100) & 20.38$\pm$0.17  \\ 
      Difficult + meaningful (DARTS-2nd, CIFAR-100) &
        \textbf{19.47}$\pm$0.20  \\
     \hline
    Difficult only (P-DARTS, CIFAR-100) &  18.12$\pm$0.11 \\ 
            Difficult + meaningful (P-DARTS, CIFAR-100) &
            \textbf{16.28}$\pm$0.10\\
         \hline
             Difficult only (DARTS-2nd, CIFAR-10) & 2.79$\pm$0.06  \\
         Difficult + meaningful (DARTS-2nd, CIFAR-10) &\textbf{2.72}$\pm$0.07   \\
         \hline
    \end{tabular} 
    \label{tab:ab1}
\end{table}

\begin{table}[t]
 \caption{
    Results for ablation setting 2. ``Test only" denotes that the tester is trained only using the created test to perform the target task. ``Test + training" denotes that the tester is trained using both the test and the training data of the target task. 
    }
    \centering
    \begin{tabular}{l|c}
    \hline
    Method & Error (\%)\\
    \hline
     Test only (DARTS-2nd, CIFAR-100) &19.81$\pm$0.06  \\ 
           Test + training  (DARTS-2nd, CIFAR-100) &   \textbf{19.47}$\pm$0.20 \\
    \hline
         Test only (P-DARTS, CIFAR-100) & 17.54$\pm$0.07 \\ 
           Test + training (P-DARTS, CIFAR-100) &\textbf{16.28}$\pm$0.10  \\
         \hline
          Test only (DARTS-2nd, CIFAR-10) & 2.75$\pm$0.03  \\
         Test + training (DARTS-2nd, CIFAR-10) &\textbf{2.72}$\pm$0.07  \\
         \hline
    \end{tabular}
    \label{tab:ab4}
\end{table}

Table~\ref{tab:ab1} shows the results for ablation setting 1. As can be seen, on both CIFAR-10 and CIFAR-100, creating tests that are both difficult and   meaningful is better than creating tests solely by maximizing difficulty. The reason is that a difficult test could be composed of bad-quality examples such as outliers and incorrectly-labeled examples. Even a highly-accurate model cannot achieve good performance on such erratic examples. To address this problem, it is necessary to make the created tests meaningful. LPT achieves meaningfulness of the tests by making the tester leverage the created tests to perform the target task. The results demonstrate that this is an effective way of improving meaningfulness.

Table~\ref{tab:ab4} shows the results for ablation setting 2. As can be seen, for both CIFAR-100 and CIFAR-10, using both the created test and the training data of the target task to train the tester performs better than using the test only. By leveraging the training data, the data encoder can be better trained. And a better encoder can help to create higher-quality tests.   

\begin{figure}[t]
    \centering
 \includegraphics[width=0.49\columnwidth]{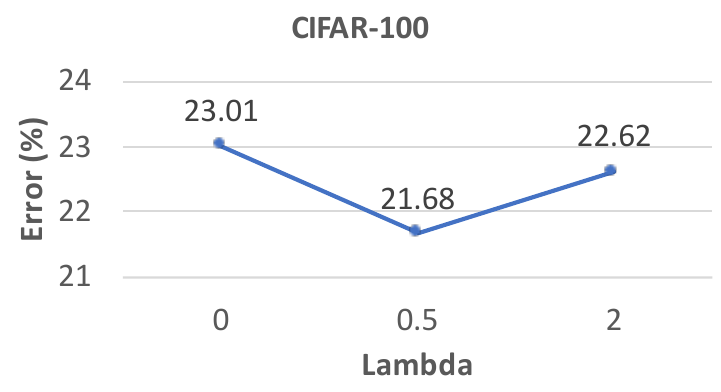}
  \includegraphics[width=0.49\columnwidth]{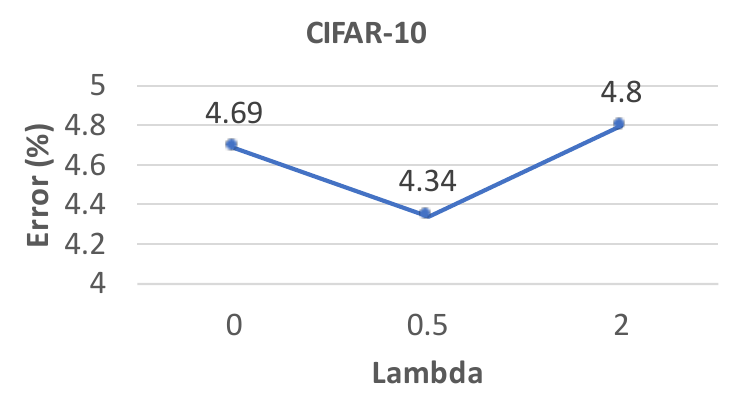}
       \caption{How errors change as $\lambda$ increases.}
 \label{fig:lambda}
\end{figure}

\begin{figure}[t]
    \centering
 \includegraphics[width=0.49\columnwidth]{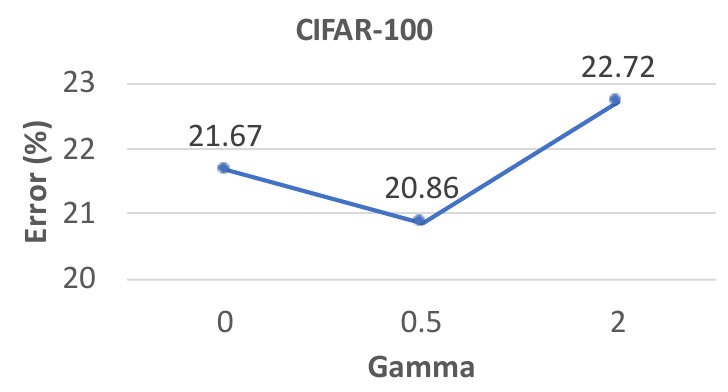}
  \includegraphics[width=0.49\columnwidth]{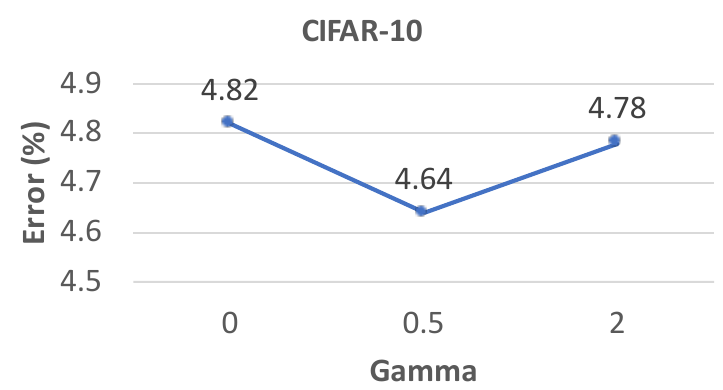}
       \caption{How errors change as $\gamma$ increases.}
 \label{fig:gamma}
\end{figure}

Figure~\ref{fig:lambda} shows  how classification errors change as $\lambda$ increases. As can be seen, on both CIFAR-100 and CIFAR-10, when $\lambda$ increases from 0 to 0.5, the error decreases. However, further increasing $\lambda$ renders the error to increase. From the tester's perspective, $\lambda$ explores a tradeoff between difficulty and meaningfulness of the tests. Increasing $\lambda$ encourages the tester to create tests that are more meaningful.  Tests with more meaningfulness can more reliably evaluate the learner. However, if $\lambda$ is too large, the tests are biased to be more meaningful but less difficult. Lacking enough difficulty, the tests may not be  compelling enough to drive the learner for improvement. Such a tradeoff effect is observed in the results on CIFAR-10 as well.

Figure~\ref{fig:gamma} shows  how classification errors change as $\gamma$ increases. As can be seen, on both CIFAR-100 and CIFAR-10, when $\gamma$ increases from 0 to 0.5, the error decreases. However, further increasing $\gamma$ renders the error to increase. Under a larger  $\gamma$, the created test plays a larger role in training the tester to perform the target task. This implicitly encourages the test creator to generate tests that are more meaningful. However, if $\gamma$ is too large,  training is dominated by the created test which incurs the following risk: if the test is not meaningful, it will result in a poor-quality data-encoder which  degrades the quality of created tests.

\section{Conclusions}
In this paper, we propose a new machine learning approach -- learning by passing tests (LPT), inspired by the test-driven learning technique of humans. In LPT, a tester model creates a sequence of tests with growing levels of difficulty. A learner model continuously improves its learning ability by striving to pass these increasingly more-challenging tests. We propose a multi-level optimization framework to formalize LPT where the tester learns to  select hard validation examples that render the learner to make large prediction errors and the learner refines its model to rectify these prediction errors. Our framework is applied for neural architecture search and achieves significant improvement on CIFAR-100, CIFAR-10, and ImageNet.

\bibliography{release-2}

\begin{thebibliography}{47}
\providecommand{\natexlab}[1]{#1}
\providecommand{\url}[1]{\texttt{#1}}
\expandafter\ifx\csname urlstyle\endcsname\relax
  \providecommand{\doi}[1]{doi: #1}\else
  \providecommand{\doi}{doi: \begingroup \urlstyle{rm}\Url}\fi

\bibitem[Bengio et~al.(2009)Bengio, Louradour, Collobert, and
  Weston]{bengio2009curriculum}
Yoshua Bengio, J{\'e}r{\^o}me Louradour, Ronan Collobert, and Jason Weston.
\newblock Curriculum learning.
\newblock In \emph{Proceedings of the 26th annual international conference on
  machine learning}, pages 41--48, 2009.

\bibitem[Cai et~al.(2019)Cai, Zhu, and Han]{cai2018proxylessnas}
Han Cai, Ligeng Zhu, and Song Han.
\newblock Proxylessnas: Direct neural architecture search on target task and
  hardware.
\newblock In \emph{{ICLR}}, 2019.

\bibitem[Casale et~al.(2019)Casale, Gordon, and Fusi]{abs-1902-05116}
Francesco~Paolo Casale, Jonathan Gordon, and Nicol{\'{o}} Fusi.
\newblock Probabilistic neural architecture search.
\newblock \emph{CoRR}, abs/1902.05116, 2019.

\bibitem[Chen and Hsieh(2020)]{abs-2002-05283}
Xiangning Chen and Cho{-}Jui Hsieh.
\newblock Stabilizing differentiable architecture search via perturbation-based
  regularization.
\newblock \emph{CoRR}, abs/2002.05283, 2020.

\bibitem[Chen et~al.(2020)Chen, Wang, Cheng, Tang, and Hsieh]{abs-2006-10355}
Xiangning Chen, Ruochen Wang, Minhao Cheng, Xiaocheng Tang, and Cho{-}Jui
  Hsieh.
\newblock Drnas: Dirichlet neural architecture search.
\newblock \emph{CoRR}, abs/2006.10355, 2020.

\bibitem[Chen et~al.(2019)Chen, Xie, Wu, and Tian]{chen2019progressive}
Xin Chen, Lingxi Xie, Jun Wu, and Qi~Tian.
\newblock Progressive differentiable architecture search: Bridging the depth
  gap between search and evaluation.
\newblock In \emph{ICCV}, 2019.

\bibitem[Chu et~al.(2019)Chu, Zhou, Zhang, and Li]{abs-1911-12126}
Xiangxiang Chu, Tianbao Zhou, Bo~Zhang, and Jixiang Li.
\newblock Fair {DARTS:} eliminating unfair advantages in differentiable
  architecture search.
\newblock \emph{CoRR}, abs/1911.12126, 2019.

\bibitem[Chu et~al.(2020{\natexlab{a}})Chu, Wang, Zhang, Lu, Wei, and
  Yan]{abs-2009-01027}
Xiangxiang Chu, Xiaoxing Wang, Bo~Zhang, Shun Lu, Xiaolin Wei, and Junchi Yan.
\newblock {DARTS-:} robustly stepping out of performance collapse without
  indicators.
\newblock \emph{CoRR}, abs/2009.01027, 2020{\natexlab{a}}.

\bibitem[Chu et~al.(2020{\natexlab{b}})Chu, Zhang, and Li]{abs-2005-03566}
Xiangxiang Chu, Bo~Zhang, and Xudong Li.
\newblock Noisy differentiable architecture search.
\newblock \emph{CoRR}, abs/2005.03566, 2020{\natexlab{b}}.

\bibitem[Deng et~al.(2009)Deng, Dong, Socher, Li, Li, and
  Fei-Fei]{deng2009imagenet}
Jia Deng, Wei Dong, Richard Socher, Li-Jia Li, Kai Li, and Li~Fei-Fei.
\newblock Imagenet: A large-scale hierarchical image database.
\newblock In \emph{2009 IEEE conference on computer vision and pattern
  recognition}, pages 248--255. Ieee, 2009.

\bibitem[Dong and Yang(2019)]{DongY19}
Xuanyi Dong and Yi~Yang.
\newblock Searching for a robust neural architecture in four {GPU} hours.
\newblock In \emph{{CVPR}}, 2019.

\bibitem[Ganin and Lempitsky(2015)]{ganin2015unsupervised}
Yaroslav Ganin and Victor Lempitsky.
\newblock Unsupervised domain adaptation by backpropagation.
\newblock In \emph{International Conference on Machine Learning}, pages
  1180--1189, 2015.

\bibitem[Goodfellow et~al.(2014{\natexlab{a}})Goodfellow, Pouget-Abadie, Mirza,
  Xu, Warde-Farley, Ozair, Courville, and Bengio]{goodfellow2014generative}
Ian Goodfellow, Jean Pouget-Abadie, Mehdi Mirza, Bing Xu, David Warde-Farley,
  Sherjil Ozair, Aaron Courville, and Yoshua Bengio.
\newblock Generative adversarial nets.
\newblock In \emph{Advances in neural information processing systems}, pages
  2672--2680, 2014{\natexlab{a}}.

\bibitem[Goodfellow et~al.(2014{\natexlab{b}})Goodfellow, Shlens, and
  Szegedy]{goodfellow2014explaining}
Ian~J Goodfellow, Jonathon Shlens, and Christian Szegedy.
\newblock Explaining and harnessing adversarial examples.
\newblock \emph{arXiv preprint arXiv:1412.6572}, 2014{\natexlab{b}}.

\bibitem[He et~al.(2016{\natexlab{a}})He, Zhang, Ren, and Sun]{he2016deep}
Kaiming He, Xiangyu Zhang, Shaoqing Ren, and Jian Sun.
\newblock Deep residual learning for image recognition.
\newblock In \emph{CVPR}, 2016{\natexlab{a}}.

\bibitem[He et~al.(2016{\natexlab{b}})He, Zhang, Ren, and Sun]{resnet}
Kaiming He, Xiangyu Zhang, Shaoqing Ren, and Jian Sun.
\newblock Deep residual learning for image recognition.
\newblock In \emph{CVPR}, 2016{\natexlab{b}}.

\bibitem[Hong et~al.(2020)Hong, Li, Zhang, Tang, Wang, Li, and
  Yu]{HongL0TWL020}
Weijun Hong, Guilin Li, Weinan Zhang, Ruiming Tang, Yunhe Wang, Zhenguo Li, and
  Yong Yu.
\newblock Dropnas: Grouped operation dropout for differentiable architecture
  search.
\newblock In \emph{{IJCAI}}, 2020.

\bibitem[Howard et~al.(2017)Howard, Zhu, Chen, Kalenichenko, Wang, Weyand,
  Andreetto, and Adam]{HowardZCKWWAA17}
Andrew~G. Howard, Menglong Zhu, Bo~Chen, Dmitry Kalenichenko, Weijun Wang,
  Tobias Weyand, Marco Andreetto, and Hartwig Adam.
\newblock Mobilenets: Efficient convolutional neural networks for mobile vision
  applications.
\newblock \emph{CoRR}, abs/1704.04861, 2017.

\bibitem[Hu et~al.(2020)Hu, Xie, Zheng, Liu, Shi, Liu, and Lin]{HuXZLSLL20}
Shoukang Hu, Sirui Xie, Hehui Zheng, Chunxiao Liu, Jianping Shi, Xunying Liu,
  and Dahua Lin.
\newblock {DSNAS:} direct neural architecture search without parameter
  retraining.
\newblock In \emph{{CVPR}}, 2020.

\bibitem[Huang et~al.(2017)Huang, Liu, van~der Maaten, and
  Weinberger]{HuangLMW17}
Gao Huang, Zhuang Liu, Laurens van~der Maaten, and Kilian~Q. Weinberger.
\newblock Densely connected convolutional networks.
\newblock In \emph{{CVPR}}, 2017.

\bibitem[Jiang et~al.(2014)Jiang, Meng, Yu, Lan, Shan, and
  Hauptmann]{jiang2014self}
Lu~Jiang, Deyu Meng, Shoou-I Yu, Zhenzhong Lan, Shiguang Shan, and Alexander
  Hauptmann.
\newblock Self-paced learning with diversity.
\newblock \emph{Advances in Neural Information Processing Systems},
  27:\penalty0 2078--2086, 2014.

\bibitem[Kingma and Ba(2014)]{adam}
Diederik Kingma and Jimmy Ba.
\newblock Adam: A method for stochastic optimization.
\newblock \emph{International Conference on Learning Representations}, 12 2014.

\bibitem[Kumar et~al.(2010)Kumar, Packer, and Koller]{kumar2010self}
M~Pawan Kumar, Benjamin Packer, and Daphne Koller.
\newblock Self-paced learning for latent variable models.
\newblock In \emph{Advances in neural information processing systems}, pages
  1189--1197, 2010.

\bibitem[Liang et~al.(2019{\natexlab{a}})Liang, Zhang, Sun, He, Huang, Zhuang,
  and Li]{abs-1909-06035}
Hanwen Liang, Shifeng Zhang, Jiacheng Sun, Xingqiu He, Weiran Huang, Kechen
  Zhuang, and Zhenguo Li.
\newblock {DARTS+:} improved differentiable architecture search with early
  stopping.
\newblock \emph{CoRR}, abs/1909.06035, 2019{\natexlab{a}}.

\bibitem[Liang et~al.(2019{\natexlab{b}})Liang, Zhang, Sun, He, Huang, Zhuang,
  and Li]{liang2019darts+}
Hanwen Liang, Shifeng Zhang, Jiacheng Sun, Xingqiu He, Weiran Huang, Kechen
  Zhuang, and Zhenguo Li.
\newblock Darts+: Improved differentiable architecture search with early
  stopping.
\newblock \emph{arXiv preprint arXiv:1909.06035}, 2019{\natexlab{b}}.

\bibitem[Liu et~al.(2018{\natexlab{a}})Liu, Zoph, Neumann, Shlens, Hua, Li,
  Fei{-}Fei, Yuille, Huang, and Murphy]{LiuZNSHLFYHM18}
Chenxi Liu, Barret Zoph, Maxim Neumann, Jonathon Shlens, Wei Hua, Li{-}Jia Li,
  Li~Fei{-}Fei, Alan~L. Yuille, Jonathan Huang, and Kevin Murphy.
\newblock Progressive neural architecture search.
\newblock In \emph{{ECCV}}, 2018{\natexlab{a}}.

\bibitem[Liu et~al.(2018{\natexlab{b}})Liu, Simonyan, Vinyals, Fernando, and
  Kavukcuoglu]{liu2017hierarchical}
Hanxiao Liu, Karen Simonyan, Oriol Vinyals, Chrisantha Fernando, and Koray
  Kavukcuoglu.
\newblock Hierarchical representations for efficient architecture search.
\newblock In \emph{{ICLR}}, 2018{\natexlab{b}}.

\bibitem[Liu et~al.(2019)Liu, Simonyan, and Yang]{liu2018darts}
Hanxiao Liu, Karen Simonyan, and Yiming Yang.
\newblock {DARTS:} differentiable architecture search.
\newblock In \emph{{ICLR}}, 2019.

\bibitem[Luo et~al.(2018)Luo, Tian, Qin, Chen, and Liu]{LuoTQCL18}
Renqian Luo, Fei Tian, Tao Qin, Enhong Chen, and Tie{-}Yan Liu.
\newblock Neural architecture optimization.
\newblock In \emph{NeurIPS}, 2018.

\bibitem[Ma et~al.(2018)Ma, Zhang, Zheng, and Sun]{MaZZS18}
Ningning Ma, Xiangyu Zhang, Hai{-}Tao Zheng, and Jian Sun.
\newblock Shufflenet {V2:} practical guidelines for efficient {CNN}
  architecture design.
\newblock In \emph{{ECCV}}, 2018.

\bibitem[Matiisen et~al.(2019)Matiisen, Oliver, Cohen, and
  Schulman]{matiisen2019teacher}
Tambet Matiisen, Avital Oliver, Taco Cohen, and John Schulman.
\newblock Teacher-student curriculum learning.
\newblock \emph{IEEE transactions on neural networks and learning systems},
  2019.

\bibitem[Noy et~al.(2020)Noy, Nayman, Ridnik, Zamir, Doveh, Friedman, Giryes,
  and Zelnik]{NoyNRZDFGZ20}
Asaf Noy, Niv Nayman, Tal Ridnik, Nadav Zamir, Sivan Doveh, Itamar Friedman,
  Raja Giryes, and Lihi Zelnik.
\newblock {ASAP:} architecture search, anneal and prune.
\newblock In \emph{{AISTATS}}, 2020.

\bibitem[Pham et~al.(2018)Pham, Guan, Zoph, Le, and Dean]{pham2018efficient}
Hieu Pham, Melody~Y. Guan, Barret Zoph, Quoc~V. Le, and Jeff Dean.
\newblock Efficient neural architecture search via parameter sharing.
\newblock In \emph{{ICML}}, 2018.

\bibitem[Real et~al.(2019)Real, Aggarwal, Huang, and Le]{real2019regularized}
Esteban Real, Alok Aggarwal, Yanping Huang, and Quoc~V Le.
\newblock Regularized evolution for image classifier architecture search.
\newblock In \emph{Proceedings of the aaai conference on artificial
  intelligence}, volume~33, pages 4780--4789, 2019.

\bibitem[Shu et~al.(2020)Shu, Liu, Qiu, and Yuille]{shu2020identifying}
Michelle Shu, Chenxi Liu, Weichao Qiu, and Alan Yuille.
\newblock Identifying model weakness with adversarial examiner.
\newblock In \emph{Proceedings of the AAAI Conference on Artificial
  Intelligence}, volume~34, pages 11998--12006, 2020.

\bibitem[Such et~al.(2019)Such, Rawal, Lehman, Stanley, and
  Clune]{abs-1912-07768}
Felipe~Petroski Such, Aditya Rawal, Joel Lehman, Kenneth~O. Stanley, and Jeff
  Clune.
\newblock Generative teaching networks: Accelerating neural architecture search
  by learning to generate synthetic training data.
\newblock \emph{CoRR}, abs/1912.07768, 2019.

\bibitem[Szegedy et~al.(2015)Szegedy, Liu, Jia, Sermanet, Reed, Anguelov,
  Erhan, Vanhoucke, and Rabinovich]{googlenet}
Christian Szegedy, Wei Liu, Yangqing Jia, Pierre Sermanet, Scott Reed, Dragomir
  Anguelov, Dumitru Erhan, Vincent Vanhoucke, and Andrew Rabinovich.
\newblock Going deeper with convolutions.
\newblock In \emph{CVPR}, 2015.

\bibitem[Tan et~al.(2019)Tan, Chen, Pang, Vasudevan, Sandler, Howard, and
  Le]{TanCPVSHL19}
Mingxing Tan, Bo~Chen, Ruoming Pang, Vijay Vasudevan, Mark Sandler, Andrew
  Howard, and Quoc~V. Le.
\newblock Mnasnet: Platform-aware neural architecture search for mobile.
\newblock In \emph{{CVPR}}, 2019.

\bibitem[Wang et~al.(2020)Wang, Xue, Yan, Yang, Hu, and Sun]{WangXYYHS20}
Xiaoxing Wang, Chao Xue, Junchi Yan, Xiaokang Yang, Yonggang Hu, and Kewei Sun.
\newblock Mergenas: Merge operations into one for differentiable architecture
  search.
\newblock In \emph{{IJCAI}}, 2020.

\bibitem[Xie et~al.(2019)Xie, Zheng, Liu, and Lin]{xie2018snas}
Sirui Xie, Hehui Zheng, Chunxiao Liu, and Liang Lin.
\newblock {SNAS:} stochastic neural architecture search.
\newblock In \emph{{ICLR}}, 2019.

\bibitem[Xu et~al.(2020)Xu, Xie, Zhang, Chen, Qi, Tian, and
  Xiong]{abs-1907-05737}
Yuhui Xu, Lingxi Xie, Xiaopeng Zhang, Xin Chen, Guo{-}Jun Qi, Qi~Tian, and
  Hongkai Xiong.
\newblock {PC-DARTS:} partial channel connections for memory-efficient
  architecture search.
\newblock In \emph{{ICLR}}, 2020.

\bibitem[Yu et~al.(2017)Yu, Zhang, Wang, and Yu]{yu2017seqgan}
Lantao Yu, Weinan Zhang, Jun Wang, and Yong Yu.
\newblock Seqgan: Sequence generative adversarial nets with policy gradient.
\newblock In \emph{AAAI}, 2017.

\bibitem[Zela et~al.(2020)Zela, Elsken, Saikia, Marrakchi, Brox, and
  Hutter]{ZelaESMBH20}
Arber Zela, Thomas Elsken, Tonmoy Saikia, Yassine Marrakchi, Thomas Brox, and
  Frank Hutter.
\newblock Understanding and robustifying differentiable architecture search.
\newblock In \emph{{ICLR}}, 2020.

\bibitem[Zhang et~al.(2018)Zhang, Zhou, Lin, and Sun]{ZhangZLS18}
Xiangyu Zhang, Xinyu Zhou, Mengxiao Lin, and Jian Sun.
\newblock Shufflenet: An extremely efficient convolutional neural network for
  mobile devices.
\newblock In \emph{{CVPR}}, 2018.

\bibitem[Zhou et~al.(2019)Zhou, Yang, Wang, and Pan]{ZhouYWP19}
Hongpeng Zhou, Minghao Yang, Jun Wang, and Wei Pan.
\newblock Bayesnas: {A} bayesian approach for neural architecture search.
\newblock In \emph{{ICML}}, 2019.

\bibitem[Zoph and Le(2017)]{zoph2016neural}
Barret Zoph and Quoc~V. Le.
\newblock Neural architecture search with reinforcement learning.
\newblock In \emph{{ICLR}}, 2017.

\bibitem[Zoph et~al.(2018)Zoph, Vasudevan, Shlens, and Le]{zoph2018learning}
Barret Zoph, Vijay Vasudevan, Jonathon Shlens, and Quoc~V Le.
\newblock Learning transferable architectures for scalable image recognition.
\newblock In \emph{CVPR}, 2018.

\end{thebibliography}

\end{document}